\documentclass[runningheads]{llncs}
\usepackage{amsmath,amssymb}
\usepackage[T1]{fontenc}
\usepackage{graphicx}
\usepackage{multirow}
\usepackage{graphicx}
\usepackage{marvosym}
\usepackage{enumerate}
\usepackage{amssymb}
\usepackage{tabularx}
\usepackage{hyperref}
\begin{document}
\title{E-BayesSAM: Efficient Bayesian Adaptation of SAM with Self-Optimizing KAN-Based Interpretation for Uncertainty-Aware Ultrasonic Segmentation}
\titlerunning{E-BayesSAM: Bayesian SAM with SO-KAN for Uncertainty Segmentation}
%
\author{Bin Huang\inst{1,3} \and
Zhong Liu\inst{1} \and
Huiying Wen\inst{2} \and
Bingsheng Huang\inst{3}\textsuperscript{(\Letter)} \and
Xin Chen\inst{1}\textsuperscript{(\Letter)} \and
Shuo Li\inst{4}
}
\authorrunning{B. Huang et al.}
%
\institute{the National-Reginoal Key Technology Engineering Laboratory for Medical Ultrasound, Guangdong Key Laboratory of Biomedical Measurements and Ultrasound Imaging, School of Biomedical Engineering, Shenzhen University Medical School, Shenzhen University, Shenzhen, China\\ \email{chenxin@szu.edu.cn}\\ \and the Institute of Maternal and Child Medicine, Shenzhen Maternity and Child Healthcare Hospital, Southern Medical University, Shenzhen, China\\ \and Medical AI Lab, School of Biomedical Engineering, Shenzhen University Medical School, Shenzhen University, Shenzhen 518000, China \\ \email{huangb@szu.edu.cn} \and the Department of Biomedical Engineering and the Department of Computer and Data Science, Case Western Reserve University, Cleveland, OH, USA}
\maketitle              
\begin{abstract}
Although the Segment Anything Model (SAM) has advanced medical image segmentation, its Bayesian adaptation for uncertainty-aware segmentation remains hindered by three key issues: (1) instability in Bayesian fine-tuning of large pre-trained SAMs; (2) high computation cost due to SAM’s massive parameters; (3) SAM's black-box design limits interpretability. To overcome these, we propose E-BayesSAM, an efficient framework combining Token-wise Variational Bayesian Inference (T-VBI) for efficienty Bayesian adaptation and Self-Optimizing Kolmogorov-Arnold Network (SO-KAN) for improving interpretability. T-VBI innovatively reinterprets SAM's output tokens as dynamic probabilistic weights and reparameterizes them as latent variables without auxiliary training, enabling training-free VBI for uncertainty estimation. SO-KAN improves token prediction with learnable spline activations via self-supervised learning, providing insight to prune redundant tokens to boost efficiency and accuracy. Experiments on five ultrasound datasets demonstrated that E-BayesSAM achieves: (i) real-time inference (0.03s/image), (ii) superior segmentation accuracy (average DSC: Pruned E-BayesSAM's 89.0\% vs. E-BayesSAM's 88.0\% vs. MedSAM’s 88.3\%), and (iii) identification of four critical tokens governing SAM’s decisions. By unifying efficiency, reliability, and interpretability, E-BayesSAM bridges SAM’s versatility with clinical needs, advancing deployment in safety-critical medical applications. The source code is available at \href{https://github.com/mp31192/E-BayesSAM}{GitHub}.

\end{abstract}
\section{Introduction}
\begin{figure}[!h]
  \centering
  \includegraphics[width=\textwidth]{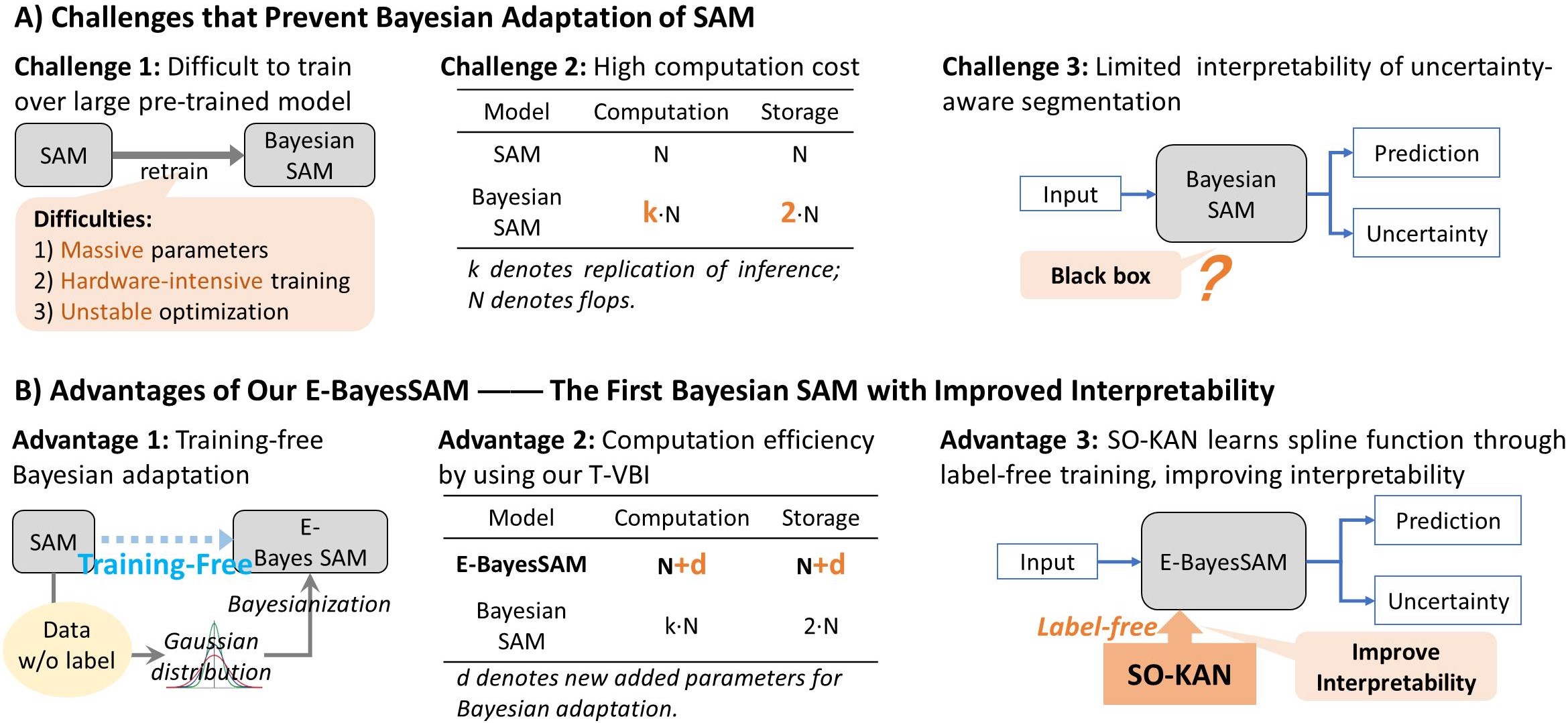}
  \caption{E-BayesSAM represents the first Bayesian adaptation of the Segment Anything Model (SAM) for uncertainty-aware segmentation, achieved through two proposed modules: Token-wise Variational Bayesian Inference (T-VBI) for efficient Bayesian adaptation and Self-Optimizing Kolmogorov-Arnold Network (SO-KAN) for improving interpretability.}
  \label{figure1}
\end{figure}

Developing a Bayesian adaptation of the Segment Anything Model (SAM) for uncertainty estimation faces critical computational and architectural challenges: its massive pre-trained parameters destabilize variational training dynamics, while its deterministic, black-box design inherently obscures uncertainty modeling\textemdash essential for clinical reliability (Fig. \ref{figure1}A). Although SAM and its variants \cite{c1,c2,c3,c4,c12,c13,c14,c16} have demonstrated outstanding performance, their lack of probabilistic uncertainty estimates limits their implementation in medical imaging, where ambiguous boundaries (e.g., tumor margins) and noisy inputs are commonplace, especially in ultrasound imaging \cite{cSAMinUS}. This gap restricts deployment in safety-critical tasks where reliable uncertainty estimation is essential to avoid misdiagnosis or overtreatment.

Current methods \cite{c6,cSAMU,cFNPCSAM,cMedSAMU} often estimate the uncertainty of SAM using test-time augmentation, which incurs high computational costs and remains a black-box approach. Meanwhile, conventional Bayesian methods \cite{c9,c10,c11} face practical challenges: full-network Bayesian adaptation may disrupt SAM's pre-trained knowledge, while reliance on labeled data for distribution modeling limits scalability in large-scale models \cite{cBD1,cBD2}.

To overcome the aforementioned challenges, we leverage the strength of Varitional Bayesian Inference (VBI) and Kolmogorov-Arnold Network (KAN) \cite{cKAN}. However, they cannot be directly applied to SAM, as both require supervised retraining\textemdash a process that risks disrupting SAM's zero-shot segmentation ability due to unstable optimization. To address this, we propose Token-wise VBI (T-VBI) and Self-Optimizing KAN (SO-KAN) to develop an Efficient Bayesian SAM (E-BayesSAM) with three key advantages (Fig. \ref{figure1}B): 1) We innovatively reformulate SAM's output tokens as dynamic Bayesian weights by reparameterizing them as Gaussian distributions: the original tokens serve as means ($\mu$), while standard deviations ($\sigma$) are inferred from unlabeled dataset token statistics, enabling training-free Bayesian adaptation (Advantage 1 in Fig. \ref{figure1}B). 2) Built upon Advantage 1, the proposed T-VBI introduces only minimal auxiliary parameters to SAM (Advantage 2 in Fig. \ref{figure1}B) while avoiding the computational overhead of full-network Bayesian methods. 3) We improve the multilayer perceptron (MLP) for token prediction with SO-KAN, using its transparent spline activations to explicitly interpret tokens and prune redundant tokens (Advantage 3 in Fig. \ref{figure1}B). Critically, spline parameters are trained via self-supervised learning on unlabeled medical images, eliminating the dependency on manual annotations.

To the best of our knowledge, E-BayesSAM is the first Bayesian SAM framework, achieving efficient uncertainty-aware segmentation and improved interpretability. Our contributions are threefold: 1) We creatively reformulate SAM’s output tokens as dynamic Gaussian-distributed weights, introducing fewer auxiliary parameters and accelerating VBI compared to conventional methods. 2) Our designed T-VBI preserves SAM's zero-shot capability while bypassing unstable retraining and annotation cost for uncertainty-aware segmentation. 3) SO-KAN learns spline-based activations trained via self-supervised learning to provide interpretable activation, which guides pruning of redundant tokens for enhanced segmentation accuracy.

\section{Methods}
\begin{figure}[!h]
  \centering
  \includegraphics[width=\textwidth]{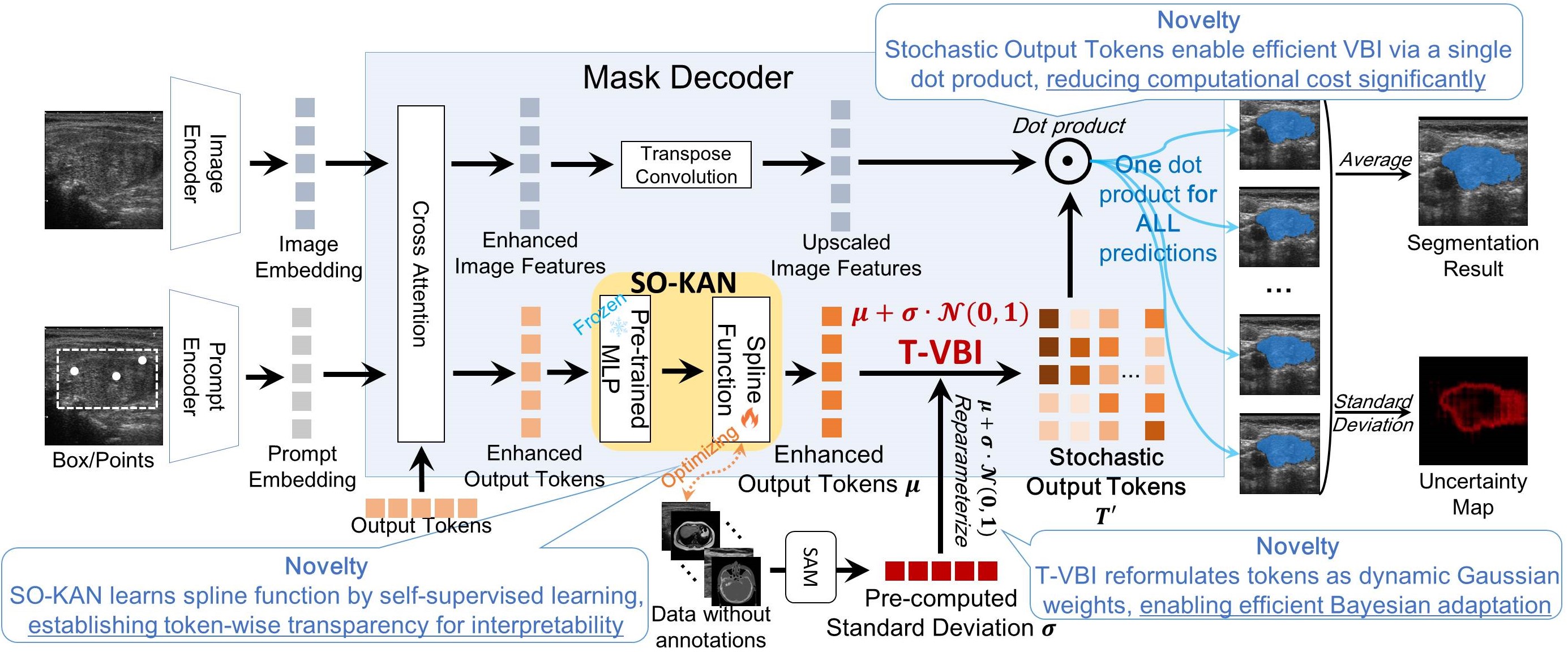}
  \caption{E-BayeSAM employs two innovations\textemdash Token-wise Varitional Bayesian Inference (T-VBI) and Self-Optimizing Kolmogorov-Arnold Network (SO-KAN)\textemdash to leverage the strengths of KAN and VBI into SAM, enabling efficient uncertainty-aware and interpretable segmentation.}
  \label{figure2}
\end{figure}

Our E-BayesSAM (Fig. \ref{figure2}) achieves Bayesian adaptation by reformulating SAM's output tokens as dynamic Bayesian weights, forming the theoretical foundation of our framework. This framework integrates two innovations\textemdash T-VBI and SO-KAN\textemdash to leverage the strengths of KAN and VBI into SAM, enabling efficient uncertainty-aware segmentation and interent interpretability without requiring manual annotations.
\subsection{Theoretical Foundation: Output Tokens are Dynamic Weights}
We creatively reinterpret the output tokens of SAM as the dynamic weights, forming the theoretical foundation of the E-BayesSAM. In conventional SAM, the segmentation mask is obtained via the point-wise product of output tokens $\mathbf{T}$ and enhanced image features $\mathbf{F}$, where $\mathbf{T}$ encodes segmentation-critical attributes (e.g., anatomical boundaries, texture patterns). Let $\mathcal{F}_\text{SAM}$ be the function class of SAM's decoder with output tokens $T\in \mathbb{R}^{k\times d}$, and $\mathcal{F}_\text{NN}$ the class of conventional neural networks with static weights $W\in \mathbb{R}^{k\times d}$. For any $f\in \mathcal{F}_{NN}$, there exists $f_\text{SAM}\in \mathcal{F}_\text{SAM}$ such that:
\begin{equation}
    \sup_{x\in\mathcal{X}}\lVert f_\text{SAM}(x)-f(x)\rVert <\epsilon
\end{equation}
where $x$ denotes the input of model, and $\epsilon >0$ depends on SAM's token generation network capacity.

SAM's tokens $T=g(x;\theta)$ are produced by a neural network $g$. By the universal approximation theorem, $g$ can approximate any continuous mapping $x\rightarrow W$ with error $\lVert g(x;\theta)-W\rVert_F<\epsilon_1$.

SAM computes the predicted masks via $M=\phi(TQ^\top)$, where $Q=\psi(F_{\text{image}})$. For a conventional network $M_\text{NN}=\phi(WQ^\top)$, the error is bounded as:
\begin{equation}
    \lVert M-M_\text{NN}\rVert_F\leq L_{\phi}\lVert TQ^\top -WQ^\top\rVert\leq L_{\phi}\lVert T-W\rVert_{F}\lVert Q\rVert_{F}
\end{equation}
Since $\lVert Q\rVert_{F}$ is bounded and $\lVert T-W\rVert_{F}<\epsilon_{1}$, the total error satisfies $\lVert M-M_\text{NN}\rVert_{F}<\epsilon$ with $\epsilon=L_{\phi}\cdot \epsilon_{1}\cdot\lVert Q\rVert_{F}$.

In conclusion, SAM's tokens $T$ emulate static weights $W$ by dynamically adapting to inputs, preserving approximation power under the assumptions of the universal approximation theorem.

\subsection{Token-wise VBI (T-VBI) for Efficient Uncertainty Estimation}
Our T-VBI reparameterizes SAM's output tokens $\mathbf{T}$ as dynamic Gaussian distributions to enable Monte Carlo-based uncertainty estimation while minimizing computational overhead. To bypass the computational inefficiency associated with Bayesian adaptation of the SAM decoder, we reformulate the output tokens $\mathbf{T}$ as dynamic probabilistic weights\textemdash the minimal subset requiring Gaussian modeling for uncertainty-aware segmentation.

The framework operates in two phases:  
1) \textbf{Standard Deviation Statistics}: Compute token-wise standard deviations $\boldsymbol{\sigma}$ from SAM's output token distribution across unlabeled training data. 2) \textbf{Reparameterized Sampling}: For new inputs $x$, derive means $\boldsymbol{\mu}$ from predicted output tokens $\mathbf{T}$ while employing $\boldsymbol{\sigma}$ pre-computed in Phase 1. The two phases are formulized as:
\begin{equation}
    \boldsymbol{\mu} = \mathbf{T} = g(x;\theta), \quad \boldsymbol{\sigma} = \sqrt{\frac{1}{N}\sum_{i=1}^N (t_{ij} - \mu_j)^2},
\end{equation}
where $\mu_j = \frac{1}{N}\sum_{i=1}^N t_{ij}$ ($t_{ij}$: $j$-th token dimension).

The latent segmentation variables (stochastic output tokens $\mathbf{T'}$) are sampled via the reparameterization trick:  
\begin{equation}
    \mathbf{T'} = \boldsymbol{\mu} + \boldsymbol{\sigma} \odot \boldsymbol{\epsilon}, \quad \boldsymbol{\epsilon} \sim \mathcal{N}(0, 1),
\end{equation}
where $\odot$ denotes element-wise multiplication, and $\mathcal{N}(0, 1)$ denotes the standard normal distribution. The segmentation mask $\hat{y}$ and uncertainty map $u$ are decoded as: 
\begin{equation}
    \hat{y} = \frac{1}{K}\sum_{i=1}^{K}\text{Sigmoid}(\mathbf{T_i'}\mathbf{Q}^\top), \quad u=\sqrt{\frac{1}{K}\sum_{i=1}^K (y_{i} - \mu_t)^2},
\end{equation}
where $\text{Sigmoid}(\cdot)$ is the sigmoid activation function, $K$ is the number of variational inference samples, and $y_{i}$ is the $i$-th prediction, and $\mu_t = \frac{1}{K}\sum_{i=1}^N y_{i}$. By confining Bayesian computation to the token space, T-VBI reduces memory overhead by 99\% compared to full-network variational inference, aligning with Fig. \ref{figure1}B.

\subsection{Self-Optimizing KAN (SO-KAN) for Interpretability and Pruning}
SO-KAN learns spline parameters via self-supervised learning while preserving SAM's zero-shot segmentation capability, providing insights to prune tokens for lightweight, uncertainty-aware segmentation. To retain SAM's pre-trained knowledge, we initialize KAN's base weight with SAM's MLP weights, freeze these weight, and train only residual spline via self-supervised learning.

To interpret and prune the tokens, spline activations are visualized to compute the positive ratio\textemdash critical since SAM's sigmoid output implies predictions rely on positive activation values. Tokens with high positive ratio (e.g., Top-4) are identified as high-contribution tokens governing boundary/texture attention for accurate segmentation; the remaining tokens are pruned to derive Pruned E-BayesSAM.  

\subsection{Model Development}
We initialize E-BayesSAM using MedSAM \cite{c2} through three core steps: (1) computing output token statistics from the SAM-MED2D \cite{c3} training set; (2) adapting SO-KAN via self-supervised learning; (3) pruning E-BayesSAM using the spline activations of KAN. For token variance estimation, per-dimension standard deviations are calculated as Equation (3). For SO-KAN training, two losses are designed: uncertainty-weighted loss $\mathcal{L}_{\text{UnW}}$ to utilize the pseudo labels and a feature-consistency loss $\mathcal{L}_{\text{feat}}$ to maintain alignment with SAM's original features. The composite loss $\mathcal{L} = \mathcal{L}_{\text{UnW}} + \mathcal{L}_{\text{feat}}$ is defined as:

\begin{equation}
    \mathcal{L}_{\text{UnW}} = \frac{1}{N}\sum_{i=1}^N \frac{1 - \mu_i}{\epsilon + \mu_i} \left\| \hat{y}_i - y^{\text{pseudo}}_i \right\|^2_2, \quad 
    \mathcal{L}_{\text{feat}} = \frac{1}{N}\sum_{i=1}^N \left\| \mathbf{T}^{\text{original}}_i - \mathbf{T}^{\text{KAN}}_i \right\|^2,
\end{equation}
where $\mu_i = \mathbb{E}[\mathbf{z}_i]$, $\epsilon=1\times 10^{-6}$, and $y^{\text{pseudo}}$ denotes conventional SAM's prediction. Data augmentation enhance data diversity via scaling (0.75x–1.25x), rotation ($\pm45^\circ$), gamma (0.5–1.5), brightness ($\pm0.1$), and Gaussian noise ($\sigma\in[0.01,0.2]$).

\subsubsection{Implementation Details} We implemented our framework in PyTorch on an Nvidia RTX 3070 GPU (8GB VRAM) with a batch size of 1, initial learning rate of 0.0001, and maximum iteration to 1,000,000, using AdamW optimizer \cite{c19}.

\section{Experimental Results and Discussion}
\subsection{Datasets and Evaluation Metrics}
We evaluate E-BayesSAM on five publicly available ultrasound datasets: two thyroid nodule datasets (DDTI \cite{cDDTI} and TN3K \cite{cTN3K}) and three breast nodule datasets (UDIAT \cite{cUDIAT}, BUSI \cite{cBUSI}, and OASBUD \cite{cOASBUD}). Segmentation accuracy is quantified using the Dice Similarity Coefficient (DSC) \cite{cDSC}.

\subsection{Compared Methods}
We benchmark E-BayesSAM against medical SAM baseline and uncertainty SAM variants under identical settings. LiteMedSAM (lightweight MedSAM) serves as the medical SAM baseline. For Pruned E-BayesSAM, we retain high-contribution tokens identified via KAN’s spline-based saliency analysis. Inference time is compared against Prompt-based augmentation \cite{cSAMU}, AutoSAM-U \cite{cAutoSAM}, and BayesFormer \cite{cBF}. All methods use box prompts generated by expanding ground-truth masks by 10 pixels to simulate annotation variability.

\subsection{Results}
\begin{table*}[!t]
\centering
\caption{Performance comparison of our E-BayesSAM with other methods in terms of DSC (\%). BayesMedSAM denotes the Bayesian MedSAM by individually using our T-VBI (Section 2.2).}
\begin{tabular}{l|ll|lll|l}
\hline
\multirow{2}{*}{Method} &
  \multicolumn{2}{l|}{Thyroid Dataset} &
  \multicolumn{3}{l|}{Breast Dataset} &
   
   \\ \cline{2-6} 
 &
  DDTI &
  TN3K &
  UDIAT &
  BUSI &
  OASBUD &
  DSC$_{Avg}$ \\ \hline
MedSAM &
  \multirow{1}{1.5cm}{90.0} &
  \multirow{1}{1.5cm}{\textbf{88.9}} &
  \multirow{1}{1.5cm}{91.0} &
  \multirow{1}{1.5cm}{\textbf{90.6}} &
  \multirow{1}{1.5cm}{80.9} &
  \multirow{1}{1.5cm}{88.3} \\
  \hline
BayesMedSAM &
  \multirow{1}{1.5cm}{89.8} &
  \multirow{1}{1.5cm}{88.8} &
  \multirow{1}{1.5cm}{90.8} &
  \multirow{1}{1.5cm}{90.5} &
  \multirow{1}{1.5cm}{80.8} &
  \multirow{1}{1.5cm}{88.1} \\
  \hline
E-BayesSAM &
  \multirow{1}{1.5cm}{89.4} &
  \multirow{1}{1.5cm}{88.7} &
  \multirow{1}{1.5cm}{90.6} &
  \multirow{1}{1.5cm}{90.5} &
  \multirow{1}{1.5cm}{80.8} &
  \multirow{1}{1.5cm}{88.0} \\
  \hline
Pruned E-BayesSAM &
  \multirow{1}{1.5cm}{\textbf{90.3}} &
  \multirow{1}{1.5cm}{88.2} &
  \multirow{1}{1.5cm}{\textbf{91.8}} &
  \multirow{1}{1.5cm}{90.2} &
  \multirow{1}{1.5cm}{\textbf{84.7}} &
  \multirow{1}{1.5cm}{\textbf{89.0}} \\
  \hline
\end{tabular}%
\label{table2}
\end{table*}
\subsubsection{Performance of E-BayesSAM}
\begin{figure}[!t]
  \centering
  \includegraphics[width=\textwidth]{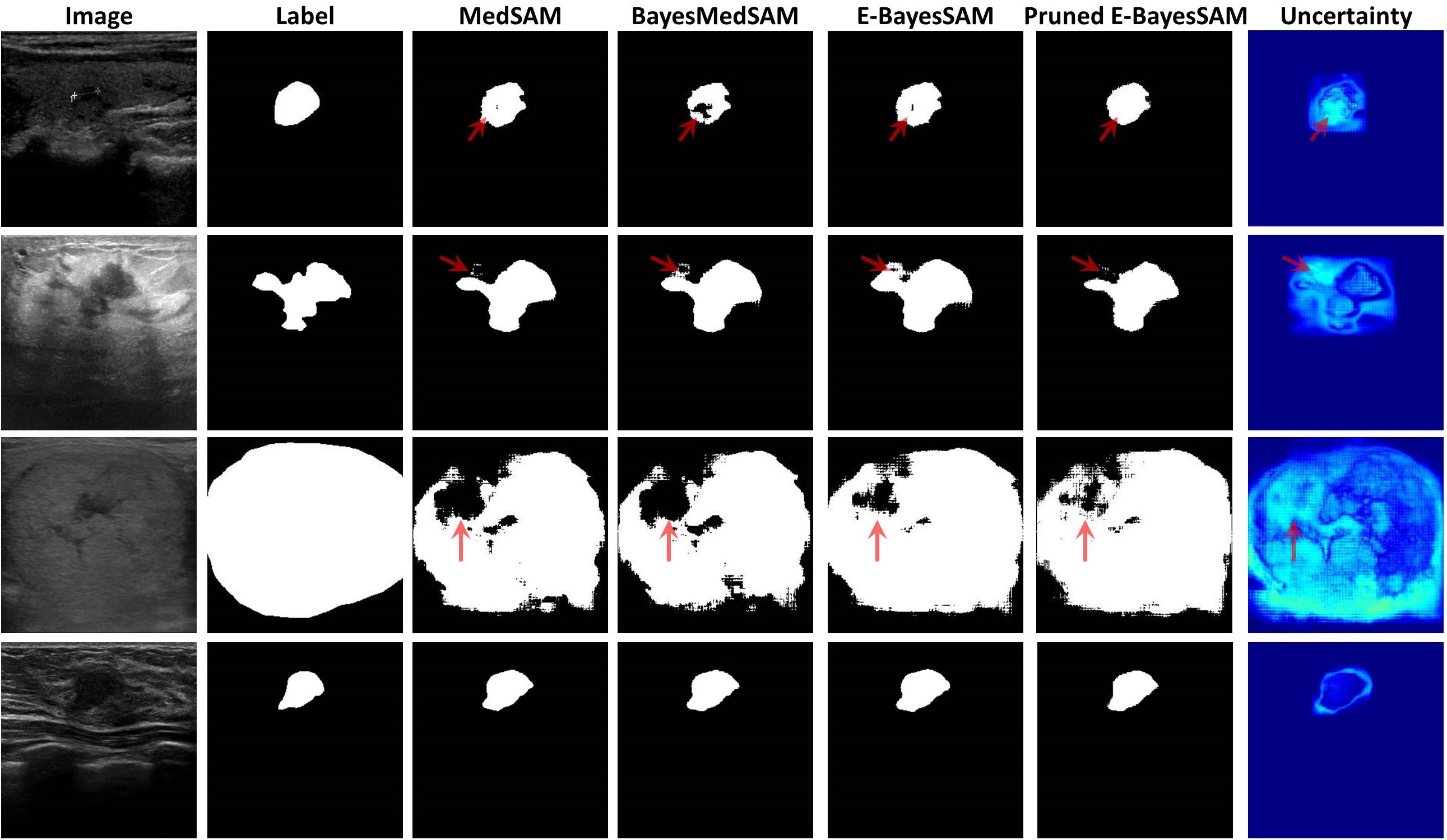}
  \caption{E-BayesSAM achieves uncertainty-aware segmentation without training, and its pruned version even shows the outstanding performance.}
  \label{figureR1}
\end{figure}
As demonstrated in Fig. \ref{figureR1}, our E-BayesSAM successfully estimates segmentation uncertainty via T-VBI while preserving SAM's zero-shot segmentation capability (Table \ref{table2}). Quantitative results show a marginal decrease in average DSC (E-BayesSAM: 88.0\% vs. MedSAM: 88.3\%) but crucially enhance prediction reliability for clinical deployment. Qualitative analysis in Fig. \ref{figureR1} reveals E-BayesSAM's dual behavior: improved segmentation of ambiguous boundaries (row 1 and 3 in Fig. \ref{figureR1}) at the cost of slightly reducing accuracy on well-defined structures (row 2 and 4 in Fig. \ref{figureR1}). We hypothesize this stems from KAN layers' targeted tokens modifications\textemdash prioritizing uncertainty-sensitive features over anatomically stable regions, aligning with clinical prioritization of challenging cases.
\begin{figure}[!t]
  \centering
  \includegraphics[width=\textwidth]{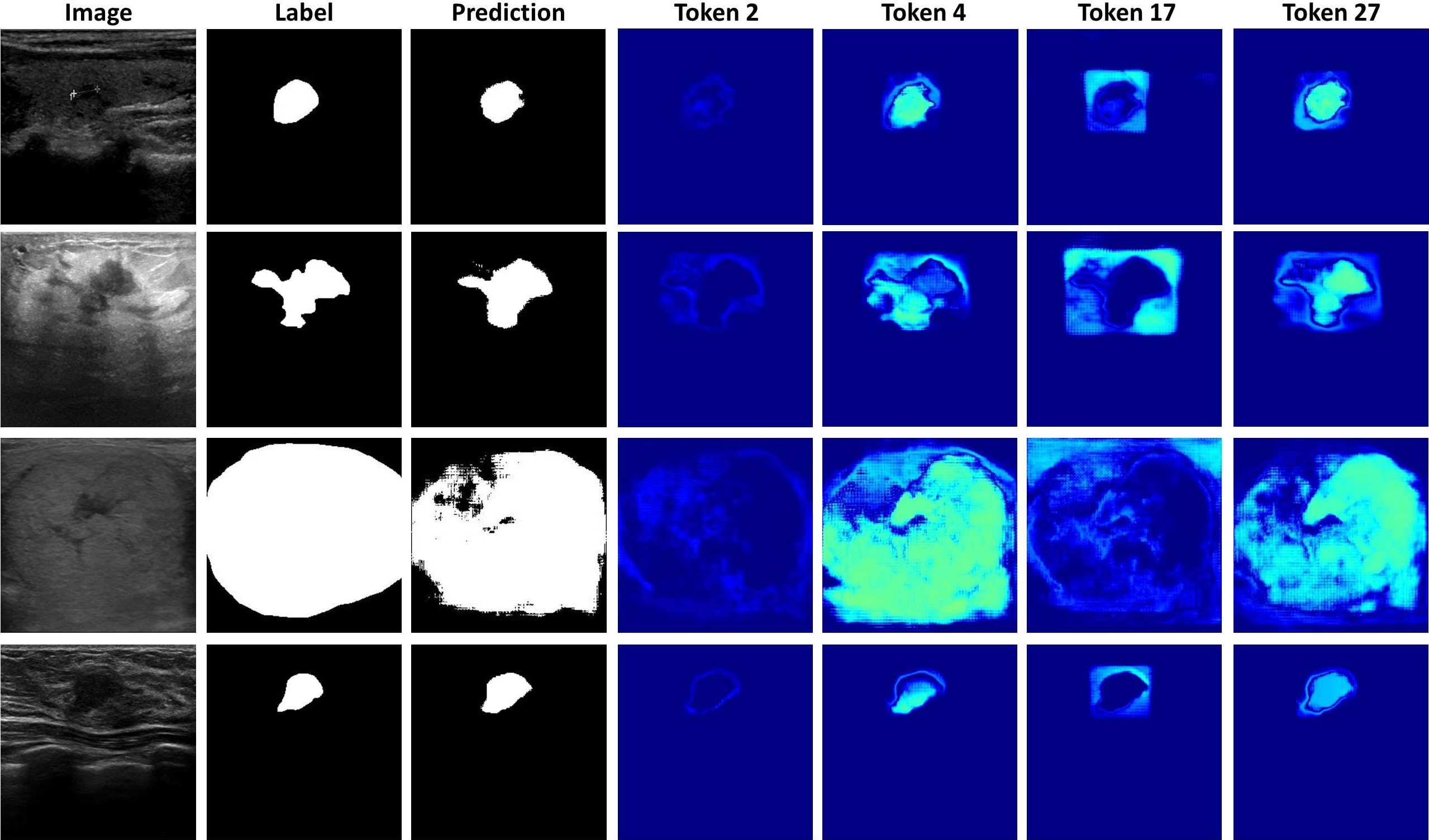}
  \caption{Four high-contribution tokens identified by KAN encode critical segmentation cues\textemdash boundary localization, foreground-background differentiation, and anatomical context\textemdash achieving accurate segmentation.}
  \label{figureR2}
\end{figure}

\subsubsection{SO-KAN for Improving Interpretability and Accuracy}
Contribution analysis of KAN activations identifies four high-contribution tokens governing SAM's segmentation decisions (Fig. \ref{figureR2}). Since retaining five tokens decreases DSC by 5\%, we select four tokens to optimize performance. These tokens correlate with anatomical boundary detection and background suppression\textemdash features essential for robust segmentation. Pruning the 28 lowest-contribution tokens reduces computation by 80\% while improving baseline DSC by 0.7\%. A similar observation was reported in DToP \cite{cDToP}, which adaptively prunes tokens in SegViT and achieves a 0.5\% improvement in mIoU.

\subsubsection{Efficiency of E-BayesSAM for Uncertainty Estimation}
E-BayesSAM achieves efficient uncertainty estimation by confining Bayesian computation to SAM's output tokens, avoiding full-network stochasticity. It processes 256$\times$256 inputs in 0.03s/sample\textemdash 23× faster than BayesFormer (0.70s) and 18× faster than prompt augmentation (0.55s). This efficiency arises from token-wise reparameterization of T-VBI, adding 256 new parameters to SAM (9.8M), resulting in a model lighter than conventional Bayesian adaptations (>13.9M). This design enables deployment on mobile ultrasound systems with <4GB VRAM.

\section{Conclusion}
We present E-BayesSAM, a novel framework that addresses the critical challenges of Bayesian adaptation of SAM\textemdash computational inefficiency and interpretability limitations\textemdash for ultrasonic image segmentation.  By integrating our designed T-VBI and SO-KAN, E-BayesSAM leverages the strengths of VBI and KAN to achieve three key advances: (1) real-time inference (0.03s/sample), (2) improved segmentation accuracy (DSC: Pruned E-BayesSAM 89.0\% vs. SAM’s 88.3\%), and (3) intrinsic interpretability via token-wise saliency mapping. Crucially, KAN’s spline visualizations reveal that four tokens govern boundary detection and background suppression, enabling targeted token pruning without performance degradation. E-BayesSAM’s lightweight design and efficient inference bridge SAM’s versatility with the reliability required in clinical practice. It also inherits MedSAM’s multi-modality generalizability, enabling direct extension to other modalities, and provides a deployable solution for safety-critical medical applications.

\begin{credits}
\subsubsection{\ackname} This study was funded by Shenzhen Basic Science Research (Key Program) (grant number JCYJ20220818095612027).

\subsubsection{\discintname}
The authors have no competing interests to declare that are relevant to the content of this article.
\end{credits}
%
%
%
%
\bibliographystyle{splncs04}
\bibliography{main}
\end{document}